\documentclass[conference]{IEEEtran}
\IEEEoverridecommandlockouts
\usepackage{cite}
\usepackage{amsmath,amssymb,amsfonts}
\usepackage{algorithmic}
\usepackage{graphicx}
\usepackage{textcomp}
\usepackage{xcolor}
\def\BibTeX{{\rm B\kern-.05em{\sc i\kern-.025em b}\kern-.08em
    T\kern-.1667em\lower.7ex\hbox{E}\kern-.125emX}}

\usepackage{latexsym}
\usepackage{url}
\usepackage{booktabs}
\usepackage{multirow}
\usepackage{caption}
\usepackage{subcaption}
\usepackage{hyperref}
\usepackage{xurl}

\begin{document}

\title{Benchmarking Deep Learning Models for Anomaly Detection in Real-World Cloud Telemetry
}

\author{\IEEEauthorblockN{Sumit Dhyani}
\IEEEauthorblockA{\textit{Electrical Engineering Department} \\
\textit{Indian Institute of Technology}\\
Delhi, India \\
eet242364@ee.iitd.ac.in }
\and
\IEEEauthorblockN{Sougata Mukherjea}
\IEEEauthorblockA{\textit{Electrical Engineering Department} \\
\textit{Indian Institute of Technology}\\
Delhi, India \\
sougatam@iitd.ac.in}
}

\maketitle

\begin{abstract}
Modern cloud platforms generate high-dimensional, noisy telemetry from thousands of microservices. Unlike classical anomaly detection benchmarks, cloud anomalies are short-lived, irregular, and largely non-periodic, making timely detection challenging. We present an empirical benchmark of deep learning models for anomaly detection using a real-world IBM Cloud Console dataset containing 39,000+ time steps, 117,000 service–feature combinations, and 25 labeled anomaly windows.

We evaluate GRU, Transformer, TimesNet, and Autoformer architectures using the Numenta Anomaly Benchmark (NAB), a timing-aware metric that accounts for early detection and false alarms. TimesNet achieves the highest NAB scores due to strong early-detection sensitivity. Transformer models also detect anomalies early but incur higher false-positive rates. An improved GRU autoencoder, incorporating robust scoring and persistence rules, achieves competitive performance with a more stable false-alarm profile. Overall, the modest performance differences suggest that real cloud anomalies are dominated by short-term deviations rather than persistent periodic structure, emphasizing the importance of timing-aware evaluation and robust post-processing.
\end{abstract}


\section{Introduction}

Modern cloud platforms operate as highly distributed ecosystems composed of thousands of microservices deployed across multiple regions and data centers. These systems generate high-dimensional and noisy telemetry streams that capture performance and reliability indicators. Detecting anomalous behavior in such environments is critical for maintaining system availability and preventing service degradation.

Unlike classical anomaly detection benchmarks, cloud anomalies are typically short-lived, irregular, and largely non-periodic. They are influenced by deployment changes, auto-scaling, transient workload bursts, and network jitter, and are often partially mitigated by self-healing mechanisms. As a result, techniques that perform well on clean, periodic industrial datasets often struggle when applied to realistic cloud telemetry.

Another key challenge is the operational cost of false positives. Frequent spurious alerts can trigger unnecessary auto-scaling actions and lead to alert fatigue among Site Reliability Engineers (SREs). Therefore, anomaly detection in cloud environments must emphasize not only accuracy but also \textit{early detection} and \textit{low false-alarm rates}. Traditional point-wise metrics such as F1-score and ROC-AUC fail to capture these requirements. We therefore adopt the \textbf{Numenta Anomaly Benchmark (NAB)}~\cite{lavin2015nab}, which explicitly rewards early detection while penalizing late detections and false positives.

In this work, we conduct a large-scale empirical benchmark of deep learning models for anomaly detection using real-world production telemetry from the IBM Cloud Console~\cite{islam2025ibmcloud}. The dataset is high-dimensional, noisy, and non-seasonal, providing a realistic testbed for cloud anomaly detection. We evaluate recurrent reconstruction-based models (GRU autoencoder)~\cite{cho2014gru}, attention-based models (Anomaly Transformer)~\cite{xu2022anomaly}, and periodicity-oriented forecasting models (TimesNet~\cite{wu2023timesnet} and Autoformer~\cite{shang2024autoformer}). We also introduce an improved GRU-based detector incorporating log-scaled reconstruction error, MAD-based normalization, and persistence--cooldown rules.

Our results show that TimesNet achieves the highest NAB scores, primarily due to strong early-detection sensitivity. Transformer-based models also detect anomalies early but incur higher false-positive rates in noisy telemetry streams. The improved GRU remains competitive while offering a more stable trade-off between early detection and false alarms. These findings suggest that real cloud anomalies are dominated by short-term deviations rather than persistent periodic structure, limiting the practical gains from increasingly complex architectures.

\textbf{Contributions.} Our contributions are as follows:
\begin{itemize}
\item A comprehensive benchmark of recurrent, attention-based, and periodicity-oriented models on large-scale real-world cloud telemetry.
\item A timing-aware evaluation framework using NAB aligned with operational cloud monitoring.
\item An improved GRU-based detection pipeline with robust scoring and post-processing.
\item Empirical evidence that higher model complexity does not necessarily yield proportionally better operational performance in noisy, non-seasonal cloud environments.
We release our evaluation models to facilitate future research in \url{https://github.com/Sumit2399/Anomaly-Detection-in-Cloud-Systems-A-Deep-Learning-Perspective}.
\end{itemize}

The remainder of this paper is organized as follows. Section II reviews related work. Section III describes the models and methodology. Section IV presents experimental results. Section V concludes the paper.

\section{Related Work}

Anomaly detection has been widely studied in time-series analysis across domains such as industrial monitoring, finance, and cyber-physical systems~\cite{Chandola2009}. However, much of the existing literature relies on structured, low-noise, and often periodic datasets that differ significantly from realistic cloud telemetry.

\subsection{Traditional Time-Series Benchmarks}

Common benchmarks for multivariate time-series anomaly detection include SMD~\cite{su2019omnianomaly} and MSL/SMAP~\cite{hundman2018}. These datasets exhibit relatively stable and periodic behavior, enabling models to detect anomalies as deviations from well-defined baselines. While useful for controlled algorithmic validation, they do not capture the high noise, non-stationarity, and scale of modern cloud systems.

\subsection{IBM Cloud Console Dataset}

The IBM Cloud Console study~\cite{islam2025ibmcloud} introduced one of the first large-scale real-world cloud telemetry datasets. The data spans 4.5 months across seven data centers and contains 39,365 time steps with 117,448 service--feature combinations aggregated at five-minute intervals. Twenty-five labeled anomaly windows derived from operational incidents are provided for evaluation.

Compared to classical benchmarks, the dataset is highly noisy, non-periodic, and extremely high-dimensional, making it a realistic and challenging testbed for cloud anomaly detection.

\subsection{Deep Learning for Time-Series Anomaly Detection}

Reconstruction-based models using LSTM and GRU architectures remain widely used for time-series anomaly detection~\cite{malhotra2015lstm,su2019omnianomaly}. These models learn representations of normal behavior and identify anomalies via reconstruction error.

Transformer-based approaches, such as the Anomaly Transformer~\cite{xu2022anomaly}, leverage self-attention to model long-range dependencies. Forecasting and periodicity-oriented models, including TimesNet~\cite{wu2023timesnet} and Autoformer~\cite{shang2024autoformer}, explicitly model seasonal and trend components and have demonstrated strong performance on structured datasets. However, their effectiveness on irregular and non-seasonal cloud telemetry remains less explored.

\subsection{Anomaly Detection in Cloud Systems}

Cloud environments introduce additional challenges due to distributed microservice architectures, dynamic scaling, and automated remediation mechanisms~\cite{Soldani2022}. The IBM Cloud study~\cite{islam2025ibmcloud} highlighted the limitations of basic GRU-based detection under such conditions. Other works have explored machine learning for cloud monitoring and anomaly detection~\cite{abdallah2024cloudsurvey}, though many rely on simplified evaluation protocols or synthetic anomalies.

\subsection{Evaluation Metrics}

Traditional evaluation metrics such as accuracy, precision, recall, F1-score, and ROC-AUC treat anomaly detection as a point-wise classification problem and ignore detection timing. In operational cloud environments, early detection and low false-alarm rates are critical. The Numenta Anomaly Benchmark (NAB)~\cite{lavin2015nab} addresses this limitation by rewarding early detections within anomaly windows while penalizing late detections and false positives. In this work, we report both the Standard and Low-FN NAB profiles.

\subsection{Positioning of This Work}

Unlike prior studies, this work provides a timing-aware benchmark of reconstruction-based, attention-based, and periodicity-oriented models on large-scale real cloud telemetry. By evaluating diverse architectures under NAB, we examine which modeling assumptions hold in noisy, non-seasonal cloud environments and demonstrate that increased model complexity does not necessarily yield proportionally better operational performance.
\section{Models and Methodology}
This section describes the deep learning models evaluated in this work and outlines how each model is adapted to the IBM Cloud Console dataset. Representative result plots are included to illustrate reconstruction behavior, anomaly detection timelines, and qualitative performance characteristics.

\subsection{GRU Autoencoder (Improved)}
The GRU (Gated Recurrent Unit) autoencoder is a reconstruction-based anomaly detector. It learns to encode and reconstruct normal telemetry patterns, and anomalies are identified as time steps with high reconstruction error.

We improved the GRU autoencoder presented in \cite{islam2025ibmcloud} by:
\begin{itemize}
    \item \textbf{Robust scoring} using logarithmic scaled mean squared error (MSE).
    \item \textbf{MAD-based normalization} (Median Absolute Deviation) to improve robustness against outliers.
    \item \textbf{Persistence and cooldown rules} to suppress transient false alarms and enforce temporal consistency.
\end{itemize}

\begin{figure}[h]
    \centering
     \begin{subfigure}{0.4\textwidth}
        \centering
        \includegraphics[width=\linewidth]{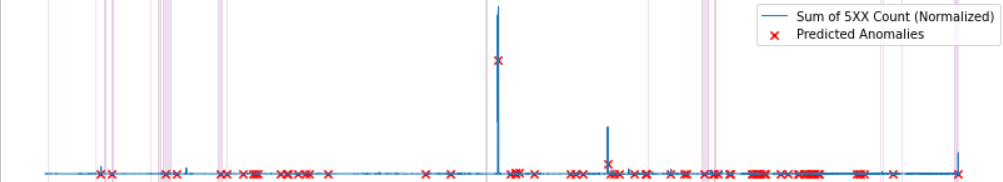}
        \caption{Sum of 5XX count (normalized) with ground truth and predicted anomalies.}
    \end{subfigure}
    \hfill
    \begin{subfigure}{0.4\textwidth}
        \centering
        \includegraphics[width=\linewidth]{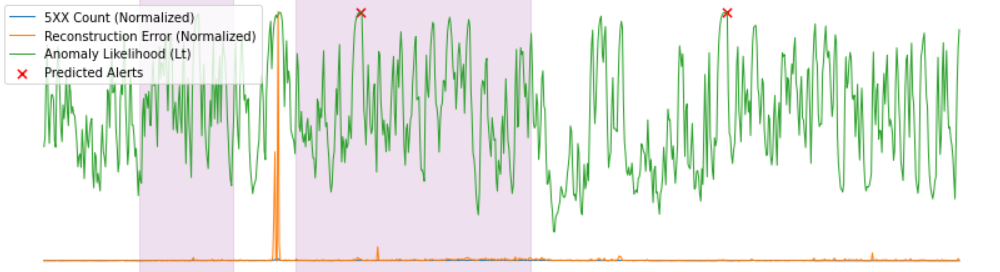}
        \caption{Reconstruction error and likelihood.}
    \end{subfigure}
    \caption{Improved GRU autoencoder results on the IBM Cloud Console dataset (Data from Mar 12--13, 2024).}
    \label{fig:gru_improved}
\end{figure}

Figure~\ref{fig:gru_improved} illustrates the anomaly detection behavior of the improved GRU autoencoder. Figure~\ref{fig:gru_improved}(a) shows how elevated 5XX error rates align with ground-truth anomaly windows and predicted alerts, demonstrating that the detector primarily triggers during sustained service failures rather than transient error spikes.
On the other hand, figure~\ref{fig:gru_improved}(b) illustrates how anomalies emerge as a consistent increase in reconstruction error and anomaly likelihood alongside rising 5XX errors, highlighting the temporal progression from raw telemetry deviation to confident anomaly detection.
\begin{itemize}
    \item During normal operation, reconstruction error remains low and stable, indicating effective learning of regular cloud telemetry patterns.
    \item Within ground-truth anomaly windows, reconstruction error increases significantly, reflecting deviations caused by sustained service disruptions and elevated 5XX errors.
    \item MAD-based normalization and persistence rules suppress short-lived fluctuations, ensuring that only persistent anomalies are detected.
\end{itemize}

\subsection{Transformer (Baseline)}
The Anomaly Transformer~\cite{xu2022anomaly} employs self-attention to model temporal dependencies across multivariate telemetry data. It is used as a reconstruction-based anomaly detector, where deviations between predicted and observed values form the anomaly score.

Despite the lack of strong periodicity in the dataset, the Transformer can capture local and medium-range dependencies, often leading to earlier anomaly detections compared to recurrent models. 

\begin{figure}[h]
    \centering
    \begin{subfigure}{0.4\textwidth}
        \centering
        \includegraphics[width=\linewidth]{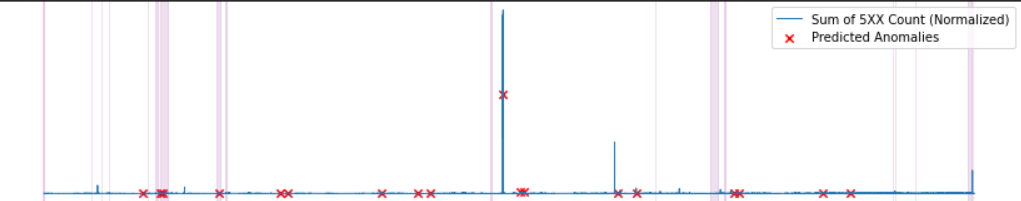}
        \caption{Sum of 5XX count (normalized) with ground truth and predicted anomalies.}
    \end{subfigure}
    \hfill
    \begin{subfigure}{0.4\textwidth}
        \centering
        \includegraphics[width=\linewidth]{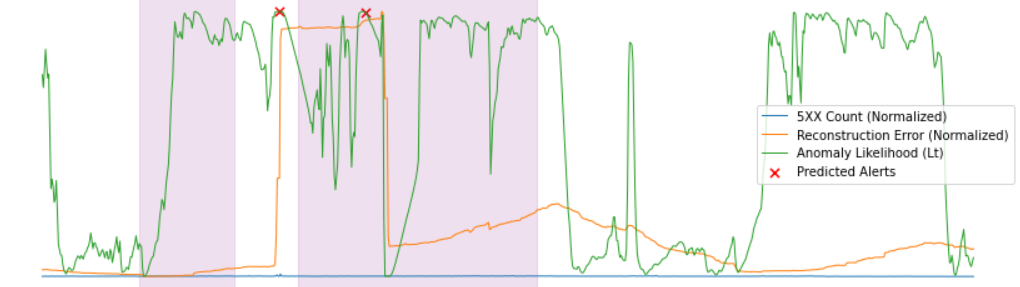}
        \caption{Reconstruction error and likelihood.}
    \end{subfigure}
    \caption{Baseline Transformer anomaly detection results (Data from Mar 12--13, 2024).}
    \label{fig:xformer_base}
\end{figure}

Figure~\ref{fig:xformer_base} shows the anomaly detection behavior of the baseline Transformer model.
\begin{itemize}
    \item The attention mechanism captures local and medium-range temporal dependencies, producing early rises in anomaly scores.
    \item Several detections appear outside of ground-truth anomaly windows, indicating sensitivity to transient telemetry noise.
    \item This suggests that reconstruction-based attention alone is insufficient to reliably distinguish noise from sustained service degradation.
\end{itemize}

\subsection{Transformer (Improved)}
The improved Transformer incorporates association-based anomaly detection concepts. Instead of relying solely on reconstruction error, it evaluates \emph{association discrepancy}, defined as deviations in learned relationships between telemetry metrics.

Key enhancements include
\begin{itemize}
    \item \textbf{Association discrepancy scoring} to capture relational changes between metrics.
    \item \textbf{Filtering of anomalous inputs} to improve model stability and reduce error propagation.
\end{itemize}

The effect of association discrepancy–based scoring are as follows:
\begin{itemize}
    \item Anomaly scores increase primarily when the relationships between telemetry metrics change, rather than due to fluctuations in isolated values.
    \item Compared to the baseline Transformer, false positives are reduced and detections are more temporally consistent.
    \item The detected anomaly windows align closely with the sustained degradation of system performance.
\end{itemize}

\subsection{TimesNet}
TimesNet~\cite{wu2023timesnet} is a temporal modeling framework designed to capture multi-periodic patterns using a two-dimensional representation and convolutional operations.Although effective for periodic datasets, IBM Cloud Console telemetry exhibits irregular and non-periodic behavior, limiting the advantage of the model. 

Despite this, TimesNet achieves the highest NAB score in our evaluation. This indicates that the model is effective at triggering early detections within anomaly windows, which is strongly rewarded by the NAB metric. However, the improvement over non-seasonal models is modest and largely comparable to Transformer-based approaches, suggesting that the weak and irregular seasonality of cloud telemetry constrains the additional benefit expected from explicit periodic modeling.

\begin{figure}[h]
    \centering
    \begin{subfigure}{0.4\textwidth}
        \centering
        \includegraphics[width=\linewidth]{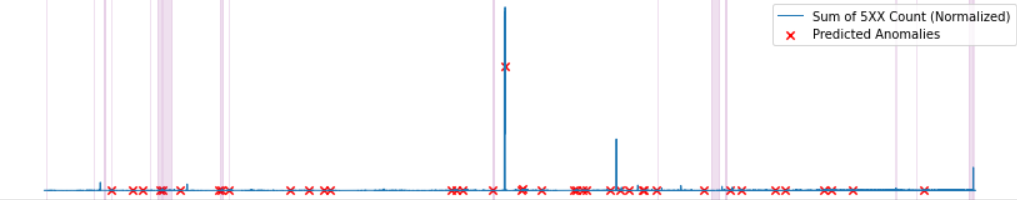}
        \caption{Sum of 5XX count (normalized) with ground truth and predicted anomalies.}
    \end{subfigure}
    \hfill
    \begin{subfigure}{0.4\textwidth}
        \centering
        \includegraphics[width=\linewidth]{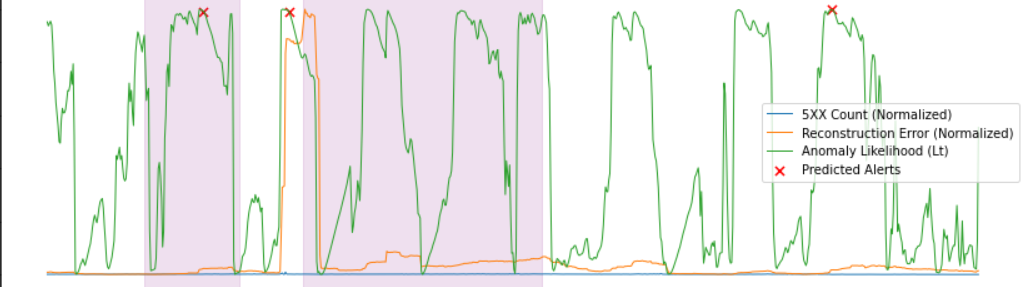}
        \caption{Reconstruction error and likelihood.}
    \end{subfigure}
    \caption{TimesNet behavior on non-seasonal cloud telemetry (Data from Mar 12--13, 2024).}
    \label{fig:timesnet}
\end{figure}

Figure~\ref{fig:timesnet} illustrates the behavior of TimesNet on cloud telemetry.
\begin{itemize}
    \item The absence of strong periodic patterns prevents a clear separation between normal and anomalous regions.
    \item Anomaly scores exhibit limited contrast, leading to missed or unstable detections.
    \item This indicates that multi-periodic modeling is less suitable for event-driven cloud environments.
\end{itemize}

\subsection{Autoformer}
Autoformer is a transformer-based forecasting model that explicitly decomposes time series into trend and seasonal components. It is often effective for long-range forecasting in datasets with strong seasonality.

In this work, Autoformer is applied as a forecasting-based anomaly detector. Deviations between predicted and observed values are used as anomaly scores. The Autoformer anomaly detection behavior is as follows:
\begin{itemize}
    \item Trend–seasonal decomposition provides limited benefit due to the non-seasonal and noisy nature of the telemetry.
    \item Forecast deviations occur during anomalies, but the anomaly boundaries are less clearly defined.
    \item Consequently, Autoformer offers only modest improvements over simpler forecasting-based approaches.
\end{itemize}

\section{Results}

\subsection{NAB scores}
\begin{table}[h]
    \centering
    \begin{tabular}{lcc}
        \toprule
        \textbf{Model} & \textbf{NAB (Standard)} & \textbf{NAB (Low-FN)} \\
        \midrule
        GRU (Baseline \cite{islam2025ibmcloud}) 
        & $17.00 \pm 0.42$ & $18.74 \pm 0.45$ \\

        GRU (Improved) 
        & $21.27 \pm 0.36$ & $23.44 \pm 0.39$ \\

        Anomaly Transformer (Baseline) 
        & $26.52 \pm 0.71$ & $28.79 \pm 0.68$ \\

        Anomaly Transformer (Improved) 
        & $26.68 \pm 0.63$ & $28.89 \pm 0.61$ \\

        TimesNet 
        & $29.05 \pm 0.28$ & $30.48 \pm 0.24$ \\

        Autoformer 
        & $26.56 \pm 0.66$ & $28.82 \pm 0.64$ \\
        \bottomrule
    \end{tabular}
    \caption{Mean and standard deviation of NAB scores over multiple runs on the IBM Cloud Console dataset.}
    \label{tab:nab_scores}
\end{table}

Table~\ref{tab:nab_scores} summarizes the NAB scores obtained by the different models under standard and low-FN profiles. Several important insights emerge from these results:

\begin{itemize}
    \item \textbf{Transformers detect anomalies earlier:}
    Both the baseline and the improved Transformer models achieve high NAB scores, indicating better timing and early detection capability.
    However, this sensitivity often leads to increased false positives due to the noisy nature of cloud telemetry.

    \item \textbf{GRU remains competitive:}
    Despite its simpler architecture, the improved GRU autoencoder achieves strong
    and stable NAB performance.
    This suggests that many cloud anomalies manifest as short-term local deviations
    rather than long-range temporal patterns.

    \item \textbf{Performance of periodicity-based models:}
    TimesNet and Autoformer are designed to exploit periodic or long-range structures. TimesNet achieves the highest NAB scores, largely due to its strong sensitivity to early deviations within anomaly windows. However, both the models
    do not consistently outperform simpler models.
    This reinforces the conclusion that cloud anomalies lack strong seasonality.

    \item \textbf{Impact of robust scoring and rules:}
    The large improvement of the GRU model compared to the baseline of the original paper \cite{islam2025ibmcloud}
 (from 6.06/14.57 to 21.27/23.44)
    highlights the importance of improved scoring functions
    and post-processing rules in practical anomaly detection.
    
    \item \textbf{Experimental Stability:}
To account for randomness arising from model initialization and training dynamics,
each experiment was repeated multiple times using different random seeds.
We report the mean and standard deviation of the NAB scores across runs.
The relatively small variance observed for all models indicates stable and reproducible
performance trends, strengthening the reliability of the reported comparisons.

\end{itemize}

\subsection{Confusion Matrix Analysis}
\subsubsection{Window-Based Confusion Matrix}

\begin{table}[h]
\centering
\begin{tabular}{lcccc}
\toprule
\textbf{Model} & \textbf{TP} & \textbf{FP} & \textbf{FN} & \textbf{TN} \\
\midrule
GRU (Baseline) & 4 & 121 & 14 & 25451 \\
GRU (Improved) & 5 & 66  & 13 & 25548 \\
Transformer (Baseline) & 6 & 163 & 12 & 25199 \\
Transformer (Improved) & 6 & 155 & 12 & 25180 \\
TimesNet & 6 & 110 & 12 & 25469 \\
Autoformer & 6 & 151 & 12 & 25215 \\
\bottomrule
\end{tabular}
\caption{Window-based confusion matrix statistics for anomaly detection.}
\label{tab:confusion_window}
\end{table}

Table~\ref{tab:confusion_window} reports the confusion matrix statistics computed
at the level of the anomaly-window, which better reflects the requirements of incident
detection in the real-world.
Transformer-based models achieve the highest number of true positives,
confirming their ability to detect most anomaly events.
However, this performance comes at the cost of increased false positives,
which can lead to alert fatigue in operational monitoring systems.

The improved GRU model exhibits a more balanced trade-off,
significantly reducing false positives while maintaining comparable
true positive detection.
This demonstrates the effectiveness of robust scoring and temporal
persistence mechanisms in stabilizing anomaly predictions.

\subsubsection{Per-Timestamp Confusion Statistics}

\begin{table}[h]
\centering
\begin{tabular}{lcccc}
\toprule
\textbf{Model} & \textbf{TP} & \textbf{FP} & \textbf{FN} & \textbf{TN} \\
\midrule
GRU (Baseline) & 8 & 163 & 865 & 25451 \\
GRU (Improved) & 5 & 66  & 868 & 25548 \\
Transformer (Baseline) & 25 & 320 & 844 & 25199 \\
Transformer (Improved) & 26 & 339 & 843 & 25180 \\
TimesNet & 8 & 114 & 865 & 25469 \\
Autoformer & 26 & 304 & 843 & 25215 \\
\bottomrule
\end{tabular}
\caption{Per-timestamp confusion matrix statistics for different models.}
\label{tab:confusion_timestamp}
\end{table}

The Per-timestamp confusion statistics in Table~\ref{tab:confusion_timestamp}
provide fine-grained insight into model sensitivity at individual time steps.
All models exhibit a large number of true negatives due to the strong class imbalance
in time-series anomaly detection.

Transformer-based models achieve higher per-timestamp true positives,
indicating responsiveness to short-lived deviations.
However, this also results in a substantial increase in false positives.
In contrast, the improved GRU model adopts a more conservative detection strategy,
trading sensitivity for improved stability and reduced spurious alerts.

\subsubsection{Summary of Error Trade-offs}

Taken together, the window-based and per-timestamp analyses reveal consistent trends.
Transformer-based models prioritize early and sensitive detection
but incur higher false alarm rates,
whereas the improved GRU model achieves a more balanced trade-off
through robust scoring and temporal consistency constraints.
These results emphasize that effective anomaly detection in cloud environments
requires not only expressive models
but also carefully designed scoring and post-processing mechanisms.

\subsection{Threshold Sensitivity and Operating Point Selection}

The performance of anomaly detection systems in operational environments is strongly influenced by the choice of detection threshold. While the Numenta Anomaly Benchmark (NAB) provides a timing-aware evaluation framework, it does not prescribe a specific threshold selection strategy. Consequently, reporting results at a single threshold without analyzing the sensitivity can lead to overly optimistic or fragile conclusions.

To address this, we conducted a systematic threshold sensitivity analysis to examine how detection performance and false alarm behavior evolve across a wide range of thresholds. This analysis serves two purposes: (i) to assess the robustness of the model under varying sensitivity levels, and (ii) to justify the selection of a principled operating point for reporting final results.

\begin{figure}[h]
    \centering
    \includegraphics[width=0.75\linewidth]{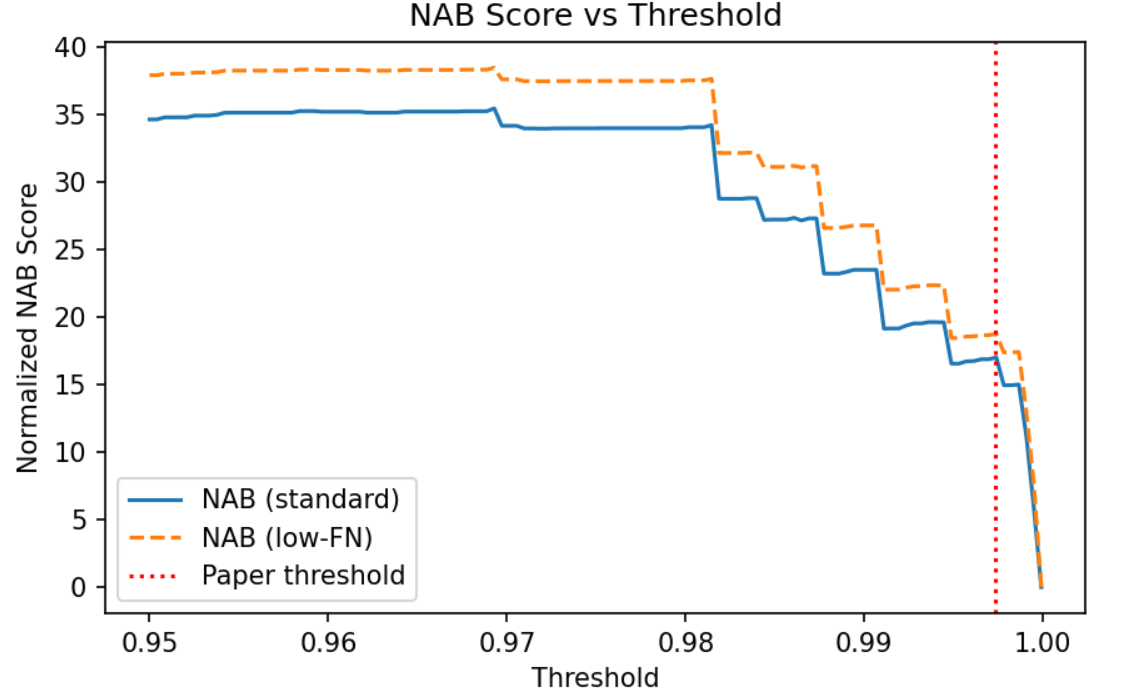}
    \caption{Normalized NAB score as a function of the anomaly detection threshold. The selected operating threshold lies within a stable region where performance remains high across small threshold variations.}
    \label{fig:nab_vs_threshold}
\end{figure}

Figure~\ref{fig:nab_vs_threshold} illustrates the variation of the normalized NAB score with respect to the detection threshold. We observe a stable plateau around the selected threshold value (0.9973), indicating that the detector’s performance is robust to minor perturbations in threshold selection. Lower thresholds can increase NAB scores by favoring early detections, but this effect is driven by a significant rise in false positive events and is therefore undesirable in practice. Excessively high thresholds, on the other hand, reduce recall and miss anomaly windows. The observed stable region suggests that strong performance is achievable without aggressive threshold tuning.

\begin{figure}[h]
    \centering
    \includegraphics[width=0.75\linewidth]{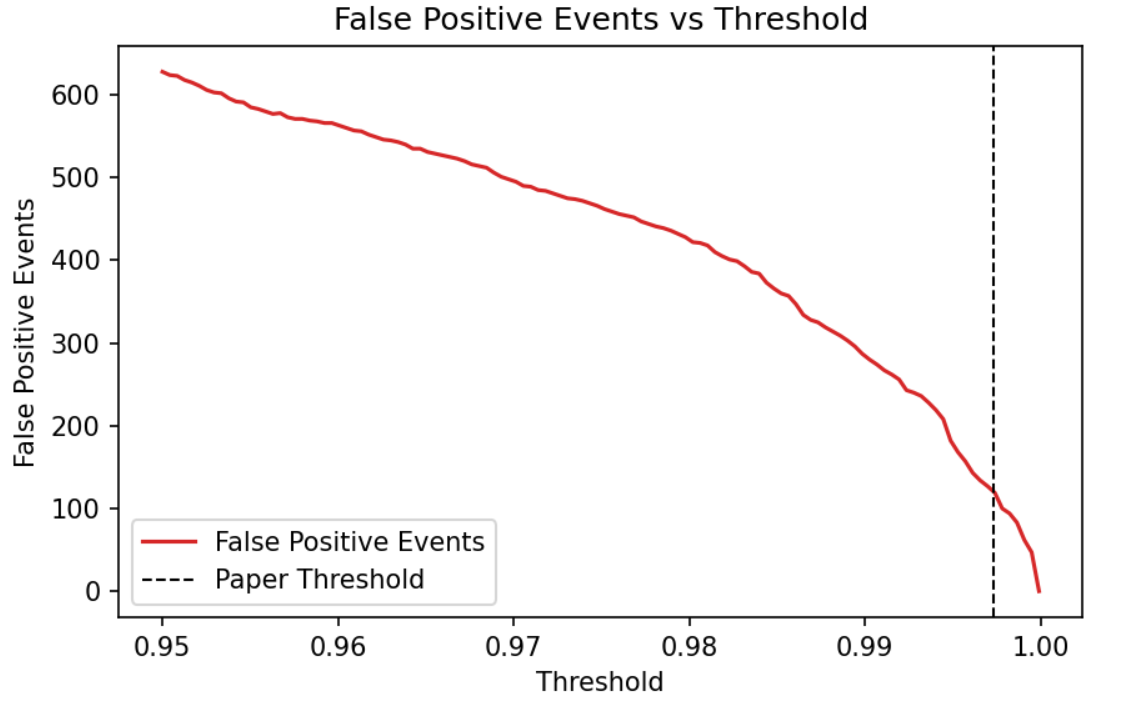}
    \caption{False positive event count as a function of the detection threshold. Lower thresholds significantly increase the number of spurious alerts, while the selected operating point maintains a low false positive rate.}
    \label{fig:fp_vs_threshold}
\end{figure}

Figure~\ref{fig:fp_vs_threshold} presents the number of false positive events as a function of the detection threshold. As expected, reducing the threshold increases the detector sensitivity but causes a sharp rise in spurious alerts. In cloud operational settings, such false positives can trigger unnecessary auto-scaling actions, increase infrastructure costs, and contribute to alert fatigue among Site Reliability Engineers (SREs). The chosen operating threshold achieves a favorable balance by maintaining a low false positive rate while preserving strong event-level detection performance.

Together, these results motivate the selection of a threshold of 0.9973 for all subsequent experiments. This operating point lies within a stable high-performance region of the NAB curve and simultaneously minimizes false positive events, aligning well with the operational priorities of large-scale cloud systems. By explicitly analyzing threshold sensitivity, we ensure that reported results are robust, reproducible, and reflective of realistic deployment constraints rather than being driven by a narrowly tuned threshold choice.

\subsection{Ablation Study of the Improved GRU Autoencoder}

To quantify the contribution of each proposed enhancement in the improved GRU autoencoder, we performed a systematic ablation study. Starting from the baseline GRU configuration, we selectively enabled or disabled the following three components:

\begin{itemize}
    \item \textbf{Log-scaled reconstruction error (Log-MSE):} applying a $\log(1 + s_t)$ transformation to stabilize large reconstruction error values.
    \item \textbf{Robust likelihood estimation:} replacing mean--variance statistics with median and MAD-based normalization when computing anomaly likelihood.
    \item \textbf{Persistence and cooldown rules:} enforcing $k$-of-$m$ persistence and a cooldown interval to suppress transient spikes and repeated alerts.
\end{itemize}

Tables~\ref{tab:gru_ablation_confusion} and~\ref{tab:gru_ablation_nab} report the per-timestamp confusion statistics and the window-based NAB scores, respectively, for different ablation settings.

\begin{table}[h]
\centering
\begin{tabular}{lcccc}
\toprule
\textbf{Configuration} & \textbf{TP} & \textbf{FP} & \textbf{FN} & \textbf{TN} \\
\midrule
Baseline (no improvements) 
& 10 & 250 & 863 & 25364 \\

Log-MSE only 
& 10 & 250 & 863 & 25364 \\

Robust likelihood only 
& 125 & 3318 & 748 & 22296 \\

Persistence only 
& 2 & 47 & 871 & 25567 \\

All components enabled
& 5 & 66 & 868 & 25548 \\
\bottomrule
\end{tabular}
\caption{Per-timestamp confusion matrix statistics for different GRU ablation configurations.}
\label{tab:gru_ablation_confusion}
\end{table}

\begin{table}[h]
\centering
\begin{tabular}{lcc}
\toprule
\textbf{Configuration} & \textbf{NAB (Standard)} & \textbf{NAB (Low-FN)} \\
\midrule
Baseline (no improvements) 
& 17.03 & 18.76 \\

Log-MSE only 
& 17.03 & 18.76 \\

Robust likelihood only 
& 41.24 & 44.16 \\

Persistence only 
& 8.48 & 9.36 \\

All components enabled
& 21.27 & 23.44 \\
\bottomrule
\end{tabular}
\caption{Window-based normalized NAB scores for different GRU ablation configurations.}
\label{tab:gru_ablation_nab}
\end{table}

\paragraph{Effect of Log-Scaled Reconstruction Error.}
Applying logarithmic scaling to the reconstruction error alone does not significantly affect detection performance. Confusion statistics and NAB scores remain close to the baseline GRU, indicating that log-scaling primarily improves numerical stability rather than anomaly separability.

\paragraph{Effect of Robust Likelihood Estimation.}
Using MAD-based likelihood estimation substantially increases the NAB score by improving sensitivity and enabling earlier detections. However, this gain comes at the cost of a sharp rise in false positives, resulting in an overly sensitive detector that is less practical for operational deployment.

\paragraph{Effect of Persistence and Cooldown Rules.}
Persistence and cooldown rules effectively suppress false positives but also reduce true anomaly detections, leading to lower recall and degraded NAB performance. This indicates that temporal smoothing must be applied to well-calibrated anomaly scores.

\paragraph{Combined Effect of All Improvements.}
When all components are combined, the improved GRU achieves a balanced trade-off between early detection and false-alarm control. Although the NAB score is slightly lower than the robust-only configuration, the false positive rate is dramatically reduced. Overall, the ablation results show that the modifications are complementary: robust likelihood improves sensitivity, persistence restores precision, and log-scaling enhances numerical stability, together yielding a practically deployable detection pipeline.
\section{Conclusion}

This paper presented a timing-aware empirical study of deep learning models for anomaly detection in real-world cloud telemetry from the IBM Cloud Console. Unlike classical benchmarks, cloud telemetry is highly noisy, non-periodic, and high-dimensional, requiring evaluation and modeling strategies aligned with operational constraints.

Using the Numenta Anomaly Benchmark (NAB), we compared GRU, Transformer, TimesNet, and Autoformer architectures under a timing-sensitive evaluation framework. TimesNet achieved the highest NAB scores due to strong early-detection sensitivity, while Transformer models also detected anomalies early, but incurred higher false-positive rates. The improved GRU model remained competitive, offering a more stable trade-off between early detection and false alarms.

Overall, this work highlights the importance of domain-specific datasets, evaluation metrics, and modeling assumptions for cloud anomaly detection. It establishes a foundation for future research on dependency-aware, streaming, and causally-informed methods capable of operating at the scale and complexity of modern cloud platforms.

\bibliographystyle{IEEEtran}
\bibliography{refs}

\end{document}